\PassOptionsToPackage{square,comma,numbers,sort&compress}{natbib}

\documentclass{article}

\usepackage[final]{neurips_2019}

\usepackage{easyReview}
\usepackage[utf8]{inputenc} % allow utf-8 input
\usepackage[T1]{fontenc}      % use 8-bit T1 fonts
\usepackage{hyperref}      		 % hyperlinks
\usepackage{url}            			% simple URL typesetting
\usepackage{nicefrac}      		  % compact symbols for 1/2, etc.
\usepackage{microtype}     		% microtypography
\usepackage[inline]{enumitem}
\usepackage{amsmath}
\usepackage{amsfonts}       	 % blackboard math symbols
\usepackage{mathtools} 
\usepackage {xcolor}
\usepackage{subcaption}
\usepackage{tabularx}
\usepackage{bm}
\usepackage{mathtools} 
\usepackage{longtable, booktabs}
\usepackage{makecell}
\usepackage{multirow}
\usepackage{diagbox}
\usepackage{latexsym}
\usepackage{stmaryrd}
\usepackage{siunitx}

\usepackage{graphicx}
\graphicspath{ {images/} }

\title{Investigating GANsformer: A Replication Study of a State-of-the-Art Image Generation Model}

\author{%
Giorgia Adorni \\
\texttt{giorgia.adorni@usi.ch} \\
\And
Felix Boelter\\
\texttt{felix.boelter@usi.ch}\\
\And
Stefano Carlo Lambertenghi\\
\texttt{stefano.carlo.lambertenghi@usi.ch}\\
}

\begin{document}

\maketitle
\begin{abstract}
The field of image generation through generative modelling is abundantly discussed nowadays. It can be used for various applications, such as up-scaling existing images, creating non-existing objects, such as interior design scenes, products or even human faces, and achieving transfer-learning processes. 
In this context, Generative Adversarial Networks (GANs) are a class of widely studied machine learning frameworks first appearing in the paper ``\emph{Generative adversarial nets}'' by \citet{goodfellow2014generative} that achieve the goal above. 
In our work, we reproduce and evaluate a novel variation of the original GAN network, the GANformer, proposed in ``\emph{Generative Adversarial Transformers}'' by \citet{hudson2021generative}. 
This project aimed to recreate the methods presented in this paper to reproduce the original results and comment on the authors’ claims. 
Due to resources and time limitations, we had to constrain the network's training times, dataset types, and sizes. 
Our research successfully recreated both variations of the proposed GANformer model and found differences between the authors’ and our results. 
Moreover, discrepancies between the publication methodology and the one implemented, made available in the code, allowed us to study two undisclosed variations of the presented procedures.
\end{abstract}

\section{Introduction}
This project investigates the reliability and reproducibility of a paper accepted for publication in a top machine learning conference. The models have been implemented using code and information provided by the authors.

With this work, we are going to verify the empirical results and claims of the paper ``\emph{Generative adversarial Transformers}'' by \citet{hudson2021generative}, by reproducing three of the computational experiments performed by the authors:
\begin{enumerate*}
    \item[(1)] the \textit{StyleGAN2} by \citet{karras2020analyzing,karras2019style}, a GAN network that uses one global latent style vector to modulate the features of each layer, hence controlling the style of all image features globally,
    \item[(2)] the \textit{GANformer} with \textit{Simplex Attention} by \citet{hudson2021generative}, which generalises the StyleGAN design with \textit{k} latent vectors that cooperate through attention. Thus, it allows for spatially finer control over the generation process since multiple style vectors impact different regions in the image concurrently, particularly permitting communication in one direction, in the generative context – from the latent vectors to the image features,
    \item[(3)] the \textit{GANformer} with \textit{Duplex Attention} by \citet{hudson2021generative}, which is based on the same principles as the previous but propagating information from global latent vectors to local image features, enabling both top-down and bottom-up reasoning to coincide.
\end{enumerate*} 

The first model is used as a baseline, while the remaining are the architectures introduced by the authors. 
They consider the GANformer as “a novel and efficient type of transformer” which demonstrates its strength and robustness over a range of tasks of visual generative modelling — simulated multi-object environments (real-world indoor and outdoor scenes) — achieving state-of-the-art results in terms of both image quality and diversity while benefiting from fast learning and better data-efficiency.

\section{Background}\label{sec:background}
\subsection{Generative Adversarial Networks (GANs)}\label{sec:gan}
Generative Adversarial Networks (GANs) \cite{goodfellow2014generative} are deep-learning-based generative models which learn to determine whether a sample is from the model or the data distribution. 
The classic architecture, illustrated in Appendix \ref{subsec:app_background} in Figure~\ref{fig:gan}, is composed of two main neural networks: the \textit{generator} ${G(z)}$, which generates new plausible examples in the problem domain — images in our case —, and the \textit{discriminator} ${D(x)}$, which classify the examples as real (coming from the training dataset) or fake (generated by $G$). 
More details are provided in Appendix~\ref{subsec:app_background}.

\subsection{StyleGAN2}\label{sec:StyleGAN}
StyleGANs are a re-design of the GANs generator architecture, which aim is to control the image synthesis process \cite{karras2019style}. The typical architecture of a StyleGAN is shown in Appendix~\ref{subsec:app_background} in Figure~\ref{fig:StyleGAN}, together with the technical details. 

In particular, we are interested in the second version of the StyleGAN, the StyleGAN2 \cite{karras2020analyzing}, which is a revisiting of the architecture of the StyleGAN synthesis network. 
Figure~\ref{fig:StyleGAN2} in Appendix \ref{subsec:app_background} exemplifies the changes made to the original architecture up to the final StyleGAN2 network. Further information is provided in Appendix \ref{subsec:app_background}.

The StyleGAN2 architecture makes it possible to control image synthesis via scale-specific style modifications. In particular, this approach attains layer-wise decomposition of visual properties, allowing StyleGAN to control global aspects of the picture, such as pose, lighting conditions or colour schemes, coherently over the entire image.
However, while this model successfully disentangles global properties, it is more limited in its ability to perform spatial decomposition. It provides no direct means to control the style of localised regions within the generated image.
\subsection{Transformers}\label{sec:transformer}
Transformers are deep-learning models based on an \textit{attention mechanism}, designed to handle sequential input data and evaluate the relationship between each input-output item \cite{vaswani2017attention}.
Unlike Recurrent Neural Networks (RNNs), this model avoids using convolutions or aligned sequences and does not necessarily require ordered input data to be processed. 
The architecture is composed by an \textit{encoder-decoder} structure where the {encoder} maps an input sequence of symbol representations $(x_1,\dots, x_n)$ to a sequence of continuous representations $z = (z_1, \dots, z_n)$, while the  {decoder}, given $z$, generates an output sequence $(y_1, \dots, y_m)$ of symbols one element at a time. 
Each transformer module (encoder-decoder) is connected via feed-forward layers.
The details about how an attention function works are reported in Appendix~\ref{subsec:app_background}. 
\citet{vaswani2017attention} used multiple multi-head attentions stacked on each other, enabling one to pass multiple input sequences simultaneously instead of one at a time, allowing for more parallelisation and reducing training times if one has access to sufficient computational resources.

\section{Methodology: Generative Adversarial Transformers}\label{sec:ganformer}
%	In this section you should give a description of the methodological aspects of your work, for 
%	instance how you modified an existing method to perform a particular task or to overcome a 
%	particular limitation. If your project is about reproducibility, here you should describe the method 
%	presented in the original paper.

The Generative Adversarial Transformers (GANsformer), introduced by \citet{hudson2021generative}, are models which combine GANs and the transformers to generate better and more realistic examples.

GANs, and in particular the StyleGAN2 model \cite{karras2020analyzing}, are used as a starting point for the GANformer design for the properties it owns as CNN: they are mighty generators for the overall style of the image since, by nature, they merge the local information of the pixels with the general information regarding the image. 
However, they are less effective for small details of localised regions within the generated image since they miss out on the long-range interaction of the faraway pixel.

Accordingly, GANsformers take advantage of the transformers' attention mechanism to make the StyleGAN2 architecture even more powerful: integrating attention in the architecture allows the network to draw global dependencies between input and output and understand the context of the image thanks to the transformer's strength for long-range interactions.
Thus, rather than focusing on global information and controlling all features globally, the transformer uses attention to propagate information from the local pixels to the global high-level representation and vice versa. 

The \textit{bipartite transformer} structure computes \textit{soft attention}, iteratively aggregating and disseminating information between the generated image features and a compact set of \textit{latent variables}, enabling bidirectional interaction between these dual representations. 
This architecture offers a solution to the StyleGAN limitation in its ability to perform spatial decomposition, which leads to the impossibility of controlling the style of a localised region within the generated image.
%	More details regarding the composition and functioning of the bipartite transformer and 
%	self-attention layers are thorough in Appendix \ref{subsec:app_methodology}. 

The \textit{transformer network} corresponds to the \textit{multi-layer bidirectional transformer encoder} (BERT), introduced by \citet{devlin2019bert}, which interleaves \textit{multi-head self-attention} and \textit{feed-forward layers}. 

The discriminator model performs multiple layers of convolution down-sampling on the image, gradually reducing the representation's resolution until making the final prediction. 
Optionally, attention can also be incorporated into the discriminator, where multiple $k$ aggregator variables use attention to collect information from the image while being processed adaptively. 

The generator likewise comprises two parts, a mapping network and a synthesis network. 
The mapping network of a GANformer is the same as that of StyleGAN2.
In the synthesis network, while in the StyleGAN2, a single global $w$ vector controls all the features equally, the GANformer uses attention so that the $k$ latent components specialise in controlling different regions in the image to create it cooperatively, and therefore perform better especially in generating images depicting multi-object scenes, also allowing for a flexible and dynamic style modulation at the region level.

\citet{hudson2021generative} have applied some adaptations to the structure of the GANformer, as presented here, to foster an interesting communication flow. The details are provided later in Section~\ref{sec:hyperparam}.
Rather than densely modelling interactions among all the pairs of pixels in the images, instead, it supports \textit{long-range adaptive interaction} between far away pixels in a moderated manner, passing through a compact and global latent bottleneck that selectively gathers information from the entire input and distributes it back to the relevant regions. 

Two attention operations could be computed over the bipartite graph, depending on the direction in which information propagates, 
\begin{enumerate*}
    \item [(1)] \textit{simplex attention} permits communication either in one way only, in the generative context, from the latent vectors to the image features, and
    \item [(2)] \textit{duplex attention}, which enables it both top-down and bottom-up.
\end{enumerate*}

\subsection{Simplex attention}
%\comment{The simplex and duplex attention layers and formulas are provided in the original article}{I'm not sure about keeping all the details regarding simplex and duplex attention since we can find these stuff exactly the same in the original paper. }
Simplex attention distributes information in a single direction over the bipartite transformer graph. 

Formally, let $X^{n\times d}$ denote an input set of $n$ vectors of dimension $d$ — where, for the image case, $n = W\times H$ — and $Y^{m\times d}$ denote a set of $m$ aggregator variables — the latent variables, in the generative case. Specifically, the attention is computed over the derived bipartite graph between these two groups of elements:
\begin{equation}
	\label{eqn:attention2}
	a(X,Y)=\mathsf{Attention}(q(X), k(Y), v(Y)) \mbox{,}
\end{equation}
where $q(\cdot), k(\cdot), v(\cdot)$ are functions that respectively map elements into queries, keys, and values, maintaining dimensionality $d$. 
The mappings are provided with positional encodings to reflect the distinct position of each element (e.g. in the image). This bipartite attention generalises self-attention, where $Y = X$.

Standard transformers implement an additive update rule of the form:
\begin{equation}
	\label{eqn:layernorm}
	u^a(X, Y)=\mathsf{LayerNorm}(X + a(X, Y)) \mbox{,}
\end{equation}
however, \cite{hudson2021generative} used the retrieved information to control both the scale as well as the bias of the elements in $X$, in line with the practice promoted by the StyleGAN model 
\cite{karras2019style}:
\begin{equation}
	\label{eqn:simplex}
	u^s(X, Y)=\gamma (a(X, Y)) \odot \omega (X) + \beta (a(X, Y)) \mbox{,}
\end{equation}
where $\gamma(\cdot), \beta(\cdot)$ are mappings that compute multiplicative and additive styles (gain and bias), maintaining dimensionality $d$, and $\omega (X) = X- \mu(X)$ normalises each element with $\sigma(X)$ respect to the other features. By normalising $X$ (image features) and then letting $Y$ (latent vectors) control the statistical tendencies of $X$, the information propagation from $Y$ to $X$ is enabled, allowing the latent vectors to control the visual generation of spatial attended regions within the image, to guide the synthesis of objects or entities.
The multiplicative integration permits significant gains in the model performance. 

\subsection{Duplex attention}
Duplex attention can be explained by taking into account the variables $Y$ to set their key-value structure: $Y = (K^{n\times d}, V^{n\times d})$, where the values store the content of the $Y$ variables, as before (e.g. the randomly sampled latent vectors in the case of GANs) while the keys track the centroids $K$ of the attention-based assignments between $Y$ and $X$, which can be computed as $K = a(Y, X)$ — namely, the weighted averages of the $X$ elements using the bipartite attention distribution derived through comparing it to $Y$. 
Consequently, the new update rule is defined as follows:
\begin{equation}
	\label{eqn:duplex}
	u^d(X, Y )=\gamma (A(X, K, V)) \odot \omega (X) + \beta (A(X, K, V)) \mbox{,}
\end{equation}
where two attention operations are compound on top of each other: first compute the \textit{soft attention} assignments between $X$ and $Y$, by $K = a(Y, X)$, and then refine the assignments by considering their centroids, by $A(X, K, V)$. This is analogous to the \textit{k-means algorithm} and works more effectively than the more straightforward update $u^a$ defined above in Equation~\eqref{eqn:simplex}.

Finally, to support bidirectional interaction between $X$ and $Y$ (the image and the latent vectors), two reciprocal simplex attentions are chained from $X$ to $Y$ and from $Y$ to $X$, obtaining the duplex attention, which alternates computing $Y:= u^a(Y, X)$ and $X:= u^d(X, Y)$, such that each representation is refined in light of its interaction with the other, integrating bottom-up and top-down interactions.

\section{Implementation}
The code from the authors has been merged with the code provided by StyleGAN2 to obtain a hybrid version of StyleGAN2 and the GANformer. In addition, we created a simplified version of the code, which removed unnecessary operations in building the network used for other model architectures. Furthermore, we implemented a Google Colab Pro version of the authors' code, as we had access to a more powerful GPU.

\subsection{Datasets}	\label{sec:dataset}
The original paper \cite{hudson2021generative} explored the GANformer model on four datasets for images and scenes: CLEVR \cite{johnson2017clevr}, LSUN-Bedrooms \cite{yu2015lsun}, Cityscapes \cite{cordts2016cityscapes} and FFHQ \cite{karras2019style}. 

Initially, we used the Cityscapes dataset since it is the smaller among the four: it contains 25k images with $256 \times 256$ resolution. 
However, the memory required to complete the training was too high on this dataset (more than 25 GB).
Even if we had more memory available, the Colab Pro's limitation to 24-hour sessions would have interrupted our experiments prematurely.

For this reason, we switch to another dataset, the Google Cartoon Set \cite{cartoonset}\footnote{	\url{https://google.github.io/cartoonset}}, containing 10k 2D cartoon avatar images with $64 \times 64$ resolution, composed of 16 components that vary in 10 artwork attributes, 4 colour attributes, and 4 proportion attributes (see Table~\ref{tab:dataset} in Appendix~\ref{sec:cartoon-results}). 

After an initial examination of the result obtained with this dataset, we decided to proceed further using a more challenging dataset, the FFHQ, exploited by the authors of the reproduced paper. 
This dataset, presented by \citet{karras2019style}, is a collection of 70k high-quality images of human faces at a $1024\times1024$ resolution, meaning that it offers much higher quality and a vastly more variation in terms of age, ethnicity, image background and coverage of accessories such as eyeglasses, sunglasses, hats, etc., than existing high-resolution datasets.

\subsection{Hyper-parameters and design choices}\label{sec:hyperparam}
In this section, we present the relevant hyper-parameters used in our experimentation, both for training and in terms of layer sizes and technical choices.

Table~\ref{tab:hyper} compares StyleGAN2 (the baseline) and the novel networks proposed in the original paper.

Note that in the code provided by the author \cite{hudson2021generative}, the hyper-parameters are not the same as mentioned in the article.

\newcolumntype{Y}{>{\raggedleft\arraybackslash}X}
\begin{table}[htb]
    \centering
    \caption{\textbf{Comparison of the hyper-parameters given in the code with those mentioned in the paper statements}. $\mbox{GANformer}_{s}$ refers to the GANformer with Simplex attention, while $\mbox{GANformer}_{d}$ refers to the GANformer with duplex attention.}
	\label{tab:hyper}
	\vspace{3mm}
	\small
	\begin{tabularx}{\linewidth}{l|rYYYY}
		\toprule
		& \textbf{StyleGAN2} & \textbf{$\mbox{GANformer}_{s}$} (code) & 
		\textbf{$\mbox{GANformer}_{d}$} (code) & \textbf{$\mbox{GANformer}_{s}$}  (article) & 
		\textbf{$\mbox{GANformer}_{d}$} (article) \\
		\midrule
		\texttt{latent\_size}    & 512    & 32   & 32    & 32    & 32 \\
		\texttt{dlatent\_size}   &512    & 32   & 32    & 32    & 32  \\
		\texttt{components\_num} & 1    & 16   & 16    & 16    & 16   \\
		\texttt{beta1}           & 0.0   & 0.0  & 0.0   & 0.9   & 0.9  \\
		\texttt{beta2}           & 0.99  & 0.99 & 0.99  & 0.999 & 0.999  \\
		\texttt{epsilon}         & 1e-8  & 1e-8 & 1e-8 &  1e-3  & 1e-3 \\
		\bottomrule                                    
	\end{tabularx}
\end{table}

A \textit{kernel size} of $k = 3$ is used after each application of the attention, together with a \textit{Leaky ReLU non-linearity} after each convolution and then up-sample or down-sample the features $X$, as part of the generator or discriminator respectively, as in StyleGAN2 \cite{karras2020analyzing}. 
To account for the features' location within the image, we use a sinusoidal positional encoding along the horizontal and vertical dimensions for the visual features $X$ and trained positional embeddings for the set of latent variables $Y$.
Overall, the bipartite transformer is thus composed of a stack that alternates attention (simplex or duplex), convolution, and up-sampling layers, starting from a $4 \times 4$ grid up to the desired resolution. 

Both the simplex and the duplex attention operations enjoy a bi-linear efficiency of $\mathcal{O}(mn)$ thanks to the network’s bipartite structure that considers all pairs of corresponding elements from $X$ and $Y$. Since, as we see below, we maintain $Y$ to be of reasonably small size, choosing m in the range of 8–32, this compares favourably to the prohibitive $\mathcal{O}(n^2)$  complexity of self-attention, which impedes its applicability to high-resolution images.

As to the loss function, optimisation and training configurations, we adopt the settings and techniques used in StyleGAN2 \cite{karras2020analyzing}, including style mixing, stochastic variation, exponential moving average for weights, and a non-saturating logistic loss with a lazy R1 regularisation.

\subsection{Experimental setup}	
The source code of our work is available at the following GitHub repository: \url{https://github.com/GiorgiaAuroraAdorni/gansformer-reproducibility-challenge}.

The approaches proposed in both the original paper codebase by \citet{karras2020analyzing} and by \citet{hudson2021generative} have been implemented in Python using TensorFlow \cite{tensorflow2015-whitepaper}, so, according to that, we used the same setup.
We created a Jupyter Notebook, which runs all the experiments in Google Colaboratory, which allows us to write and execute Python in the browser. 

All the models have been trained on a Tesla P100-PCIE-16GB (GPU) provided by Google Colab Pro.

\subsection{Computational requirements}\label{sec:comput_req}

In the original paper, \cite{hudson2021generative}, they evaluate all models under comparable conditions of the training scheme, model size, and optimisation details, implementing all the models within the codebase introduced by the StyleGAN authors \cite{karras2020analyzing}. 
All models have been trained with images of $256\times 256$  resolution and for the same number of training steps, roughly spanning a week on 2 NVIDIA V100 GPUs per model (or equivalently 3-4 days using 4 GPUs). 

Considering that we had only one GPU and not enough time to reproduce this setting, we decided to resize the images from $256\times 256$ to $64\times64$ resolution for the Google Carton Set and to $128 \times 128$ for the FFHQ dataset.

For the GANformer, we select $k = 32$ latent variables. 

All models have been trained for the same number of steps, 300 000 image training samples (300 kimg), while the paper presents results after training 100, 200, 500, 1000, 2000, 5000 and 10000 kimg samples.

For the StyleGAN2 model, we present results after training 300 kimg, obtaining good results.
Note that the original StyleGAN2 model has been trained by its authors \cite{karras2020analyzing} for up to 70000 kimg samples, which is expected to take over 90 GPU-days for a single model.

For the GANformer, the authors \cite{karras2020analyzing} show impressive results, especially when using duplex attention: the model learns much faster than competing approaches, generating astonishing images early in training. This model is expected to take 4 GPU days.

However, we are not able to replicate this achievement, first because this model learns significantly slower than the StyleGAN2, which can train approximately 1.3 times faster than the GANformer in terms of time per kimg (in the paper, they reach better results with the GANformer with 3-times less training steps than the 
StyleGAN2, but they don't specify the time required for a step).
Secondly, the GANformer with simplex attention seems to be as slow, if not slower, in achieving qualitative results regarding training steps compared to StyleGAN2. If Duplex attention is selected, qualitative results are never obtained.

As previously mentioned, we trained using Colab Pro, which enabled us to access a Tesla P100 GPU by Nvidia with 16 GB of RAM and 25 GB of RAM.

For StyleGAN2, with the given resources, for 300 kimg, training took around 8 hours, while for all variations of the GANformer, training took about 10 hours.

\section{Results}\label{sec:results}
%\comment{}{The all section should be revised since we are using also another dataset}

This section shows and comments on our results while comparing them to the original GANformer paper. We evaluate the two presented variations of the GANformer and StyleGAN2 over four metrics: Frechet Inception Distance (FID), Inception Score (IS), Precision and Recall. The FID is one of the most popular metrics for evaluating GANs, providing stable and reliable image fidelity and diversity indications. It is a measure of similarity between curves that considers the location and ordering of the points along them. FID is used to measure the feature distance between the real and the generated images for this specific application. However, it can also measure the distance between two distributions.
For this reason, we have decided to use it as a reference metric for all the following analyses. 

\section{Google Cartoon Set results}\label{sec:cartoon_results}
Starting from the Google Cartoon Set, in Table~\ref{tab:our-results}, we compared the GANformer (Simplex and Duplex) with the competing StyleGAN2 model. 
The three image synthesis methods are run for the same amount of iterations and the same dataset to have a fair comparison. To express the improvement over StyleGAN2, a difference factor of FID is positioned alongside the scores. 
\begin{table}[htb]
    \centering
    \caption{\textbf{Comparison between the GANformer (Simplex and Duplex) and competing StyleGAN2}. The last column reports the percentage of improvement for all the models, in terms of FID score, with respect to the baseline StyleGAN2 architecture.}
	\label{tab:our-results}
	\vspace{3mm}
	\small
	\begin{tabular}{l|rrrrr}
		\toprule
		\textbf{Model}  & \textbf{FID $\downarrow$}  & \textbf{IS $\uparrow$} & 
		\textbf{Precision$\uparrow$}  & \textbf{Recall $\uparrow$} & \textbf{FID Improvement (\%)}\\ 
		\midrule
		StyleGAN2                    				&  \textbf{24.77} & 2.50 & \textbf{0.0018} & \textbf{0.0211} & 0 \%\\ 
		GANformer, Simplex attention & 28.11 & \textbf{2.58} & 0.0015 & 0.0076 & -13.48 \%\\ 
		GANformer, Duplex attention  & 27.08 & 2.47 & \textbf{0.0018} & 0.0090 & -9.33  \%\\ 
		\bottomrule
	\end{tabular}
\end{table}

Unexpectedly, the novel GANformer with Duplex attention is worse than the baseline on all aspects, except the precision metric, with a staggering -9.33 \% deterioration in FID score.
To investigate this further, a similar representation is recreated in Table~\ref{tab:orig-results} using the original paper findings for the same metrics and with a mean of the scores spanning the four used datasets.
\begin{table}[htb]
    \centering
    \caption{\textbf{Original paper's reported results (mean of results over the four datasets used by the authors)}. In the last column, the percentage of improvement in all the models, in terms of FID score, is reported concerning the baseline StyleGAN2 architecture.}
	\label{tab:orig-results}
	\vspace{3mm}
	\small
	\begin{tabular}{l|rrrrr}
		\toprule
		\textbf{Model}  & \textbf{FID $\downarrow$}  & \textbf{IS $\uparrow$} & 
		\textbf{Precision$\uparrow$}  & \textbf{Recall $\uparrow$} & \textbf{FID Improvement (\%)}\\ 
		\midrule
		StyleGAN2                    & 11.29 & 2.74 & 52.02      & 23.98 & 0 \% \\ 
		{GANformer, Simplex attention} & 10.29 & \textbf{2.82}   & \textbf{56.76}     & 18.21  & 
		+8.86 \% \\ 
		{GANformer, Duplex attention}  & \textbf{7.22}   & 2.78 & 55.45     & \textbf{33.94} & 
		\textbf{+36.11 \%} 
		\\ 
		\bottomrule
	\end{tabular}
\end{table}

Notably, in the code given by the authors, attention seems to be only optionally used in the discriminator. When analysing the pre-trained models provided, it is never used, prompting us to implement two variations of the GANformer not openly discussed in \cite{hudson2021generative}.
We use the GANformer paradigm for the generator and a vanilla StyleGAN2 discriminator and obtain the results visible in Table~\ref{tab:our-results2}.
\begin{table}[htb]
    \centering
    \caption{\textbf{Comparison between the GANformer (Simplex and Duplex) both with and without attention on the Discriminator and competing StyleGAN2}. In the last column, is reported the percentage of improvement all the models, in terms of FID score, with respect to the baseline StyleGAN2 architecture.}
	\label{tab:our-results2}
	\vspace{3mm}
	\small
	\begin{tabular}{l|rrrrr}
		\toprule
		\textbf{Model}  & \textbf{FID $\downarrow$}  & \textbf{IS $\uparrow$} & 
		\textbf{Precision$\uparrow$}  & \textbf{Recall $\uparrow$} & \textbf{FID Improvement (\%)}\\ 
		\midrule
		StyleGAN2                    				&  24.77 & 2.50 & 0.0018 & 0.0211 & 0 \%\\ 
		GANformer, Simplex attention & 28.11 & 2.58 & 0.0015 & 0.0076 & -13.48 \%\\ 
		GANformer, Duplex attention  & 27.08 & 2.47 & 0.0018 & 0.0090 & -9.33  \%\\ 
		GANformer, Simplex attention (StyleGAN2 disc.) & \textbf{19.09} &  \textbf{2.62}  &  
		\textbf{0.0035}    & \textbf{0.0476}  & \textbf{+22.93 \%} \\ 
		GANformer, Duplex attention (StyleGAN2 disc.)  &  24.81  & \textbf{2.62} &   \textbf{0.0035}   
		& 0.0211 & 
		-0.16 \%\\ 
		\bottomrule
	\end{tabular}
\end{table}

\begin{figure*}[htpb]				
	\centering
%	\raggedleft
	\includegraphics[width=.6\linewidth]{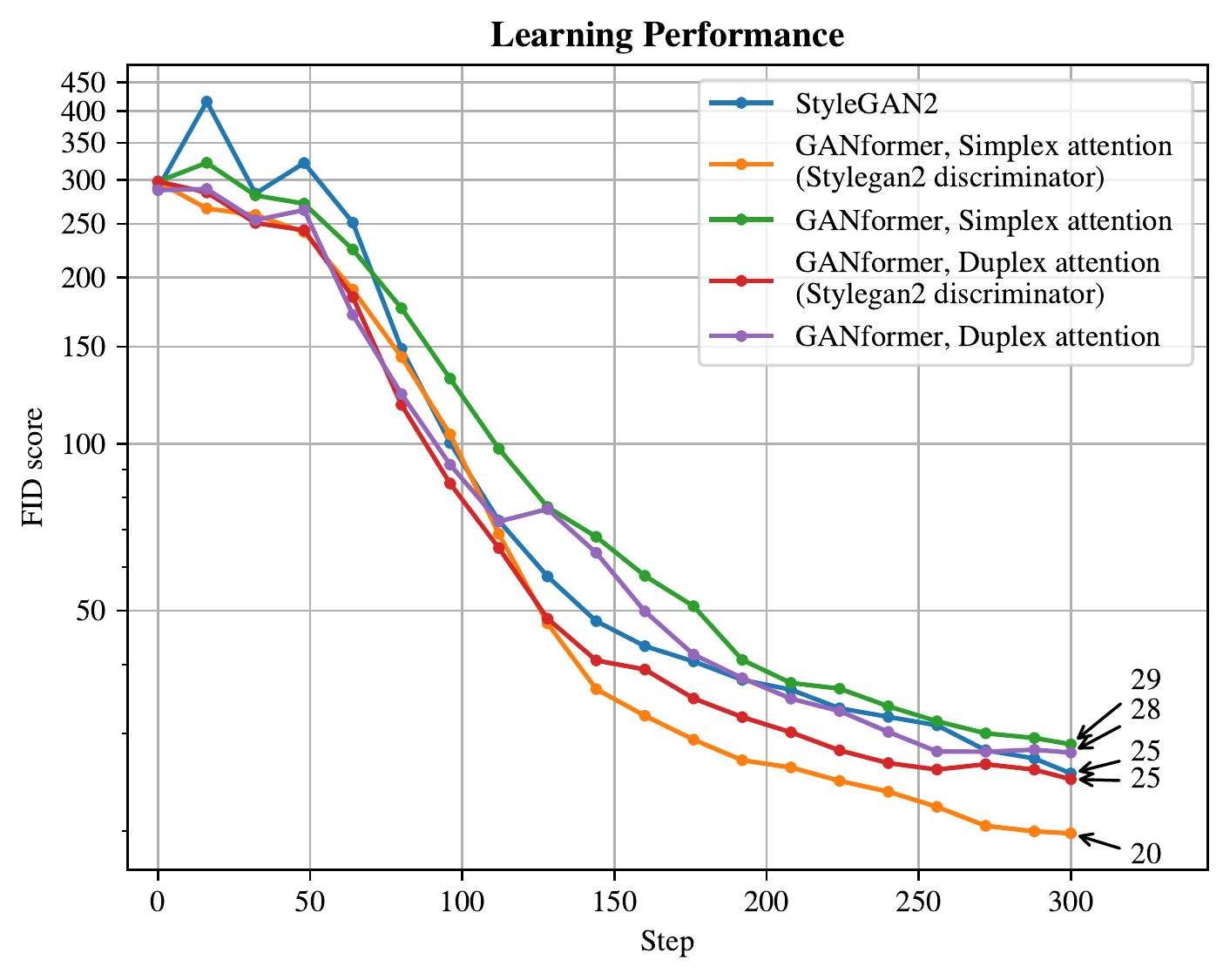}
	\caption{\textbf{Comparison between the StyleGAN2 and the GANformer models.} We evaluate the models using the FID score along 300 kimg samples. The score is computed every sixteen checkpoints.}
	\label{fig:performance}
\end{figure*}

Due to the relevance of the FID metric, we compare all the created models' scores over iterations numbers to show the results' quality over training steps in Figure~\ref{fig:performance}.

As mentioned before, unexpectedly, the model that yields better results in the original paper is not the best in our experiments.
Not only the FID score is worse than the baseline StyleGAN2 at the final training step, both for simplex and duplex attention implementations, but also claims about efficiency cannot be verified.
Models which include attention on the generator only, however, are faster in terms of steps to reach a qualitative result when compared to the baseline. 
We believe that this behaviour explains the choices made by the authors in their GitHub code publication.
Moreover, a comment has to be made on the claim of efficiency: both with and without attention to the discriminator, a training step is considerably slower to be completed on the same resources when compared to StyleGAN2.
While the latter can have a training speed of 10.9 images generated a second, all flavours of the GANformer, in the best case, only yield 8.3 images generated a second.

\begin{figure}[htb]
	\centering
	\includegraphics[width=.9\linewidth]{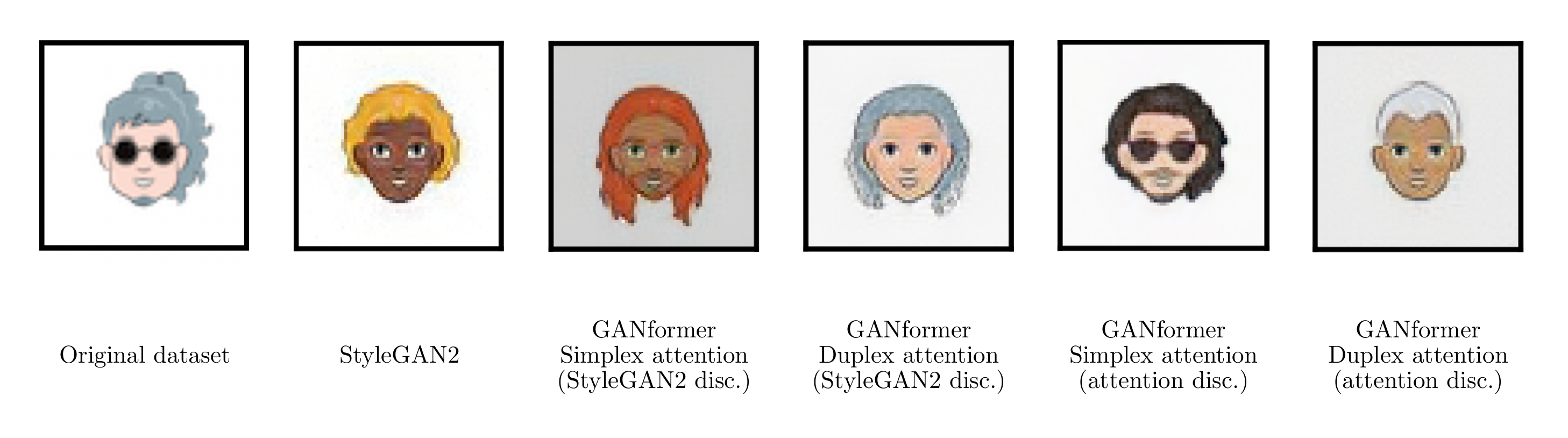}
	\caption{\textbf{Images generated using of the various models}.} 
	\label{fig:summaryImgs}
\end{figure}
Figure~\ref{fig:summaryImgs} shows an image generated by each variation.
Here, the Simplex GANformer implementation with StyleGAN2 is the model that performs better, while we couldn't reproduce the same results as the original paper for the Duplex GANformer implementation. 
In general, all the implementations can produce visually compelling results and compare all outputs to the actual images in the original dataset. All models tend to create similar results, with only background colour differing for models with attention on the generator.

Finally, to visualise our findings less empirically, we have used random seeds to create latent inputs and shown the resulting images generated by the baseline StyleGAN2 and all four presented variations of the GANformer in Figures \ref{fig:random} and \ref{fig:interpolation} in Appendix \ref{sec:cartoon-results}.

\section{FFHQ dataset results}\label{sec:ffhq_results}
For the FFHQ dataset, in Table~\ref{tab:results_ffhq}, we compared the two GANformer models, with Duplex attention on the generator and one with attention on the discriminator, too the other with a vanilla StyleGAN2 discriminator.
The three image synthesis methods are run for the same amount of iterations to have a fair comparison.
\begin{table}[htb]
    \centering
    \caption{\textbf{Comparison between the GANformer (Simplex and Duplex) both with and without attention on the discriminator and competing StyleGAN2}. The last column reports the percentage of improvement for all the models, in terms of FID score, with respect to the baseline StyleGAN2 architecture.}
	\label{tab:results_ffhq}
	\vspace{3mm}
	\small
	\begin{tabular}{l|rrrrr}
		\toprule
		\textbf{Model}  & \textbf{FID $\downarrow$}  & \textbf{IS $\uparrow$} & 
		\textbf{Precision$\uparrow$}  & \textbf{Recall $\uparrow$} & \textbf{FID Improvement (\%)}\\ 
		\midrule
		StyleGAN2                    				&  {40.22} &  3.21 &  \textbf{0.55}&  \textbf{0.0301} & 0 \%\\ 
		GANformer, Simplex attention & 45.71 & 3.35 &  \textbf{0.55}  &  0.0055 & -13.65 \%\\ 
		GANformer, Duplex attention  & {53.48} & 3.35 & 0.48& 0.0033 & -32.97 \%\\ 
		GANformer, Simplex attention (StyleGAN2 disc.)  &   \textbf{39.73}  & 3.49  & 0.53  &  0.0079 &  \textbf{+1.22} \%\\ 
		GANformer, Duplex attention (StyleGAN2 disc.)  &  {43.66}  & \textbf{3.61} &   \textbf{0.55}   & {0.0078} & -{8.55} \%\\ 
		\bottomrule
	\end{tabular}
\end{table}

Once again, we can see that the model with multiple attention is considerably worse than the model with attention only in the generator. 

\begin{figure*}[htpb]				
    \centering
%	\raggedleft
    \includegraphics[width=.6\linewidth]{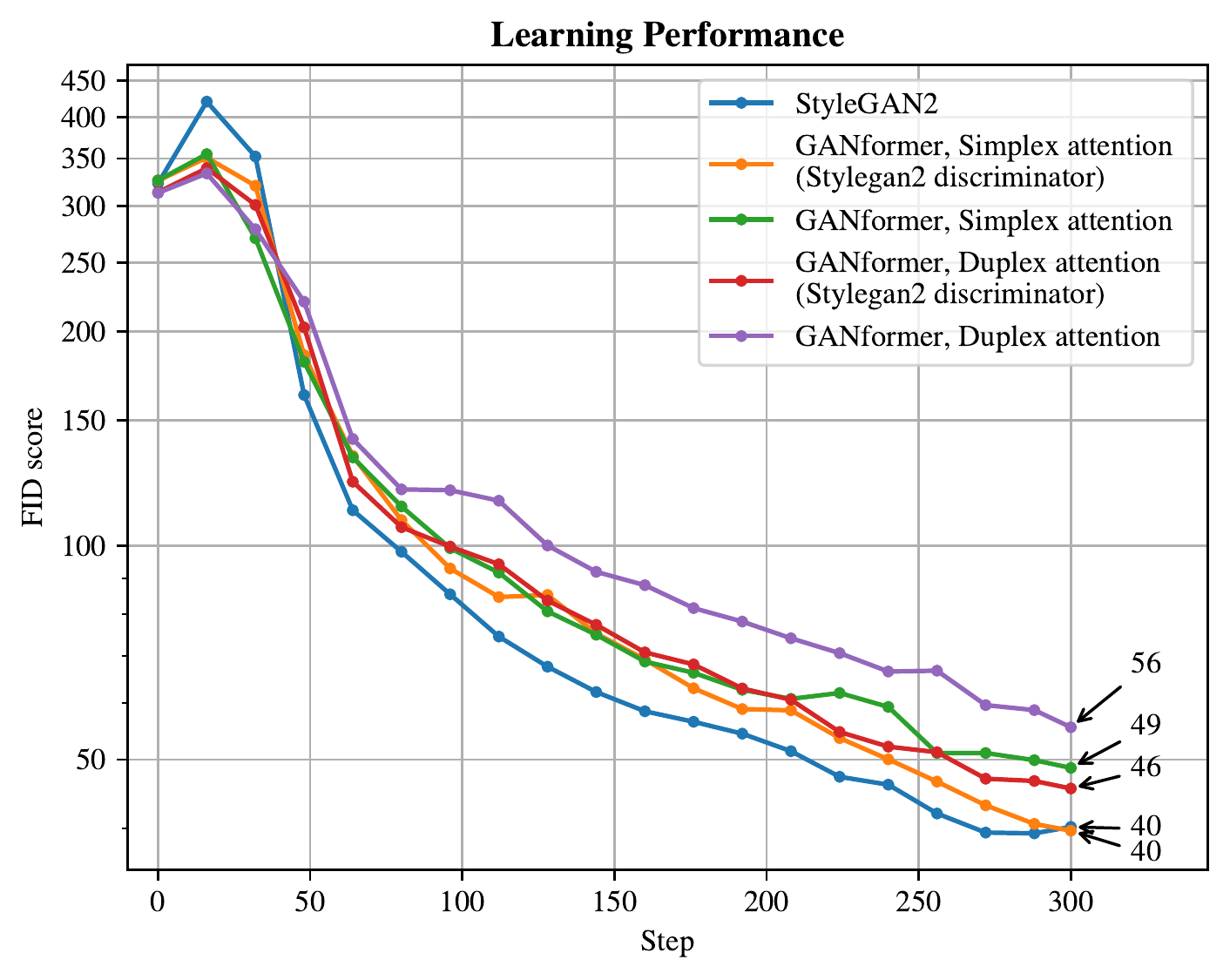}
    \caption{\textbf{Comparison between the StyleGAN2 and the GANformer models.} We evaluate the models using the FID score along 300 kimg samples. The score is computed every sixteen checkpoints.}
	\label{fig:performance-ffhq}
\end{figure*}

\begin{figure}[htb]
	\centering
	\includegraphics[width=.9\linewidth]{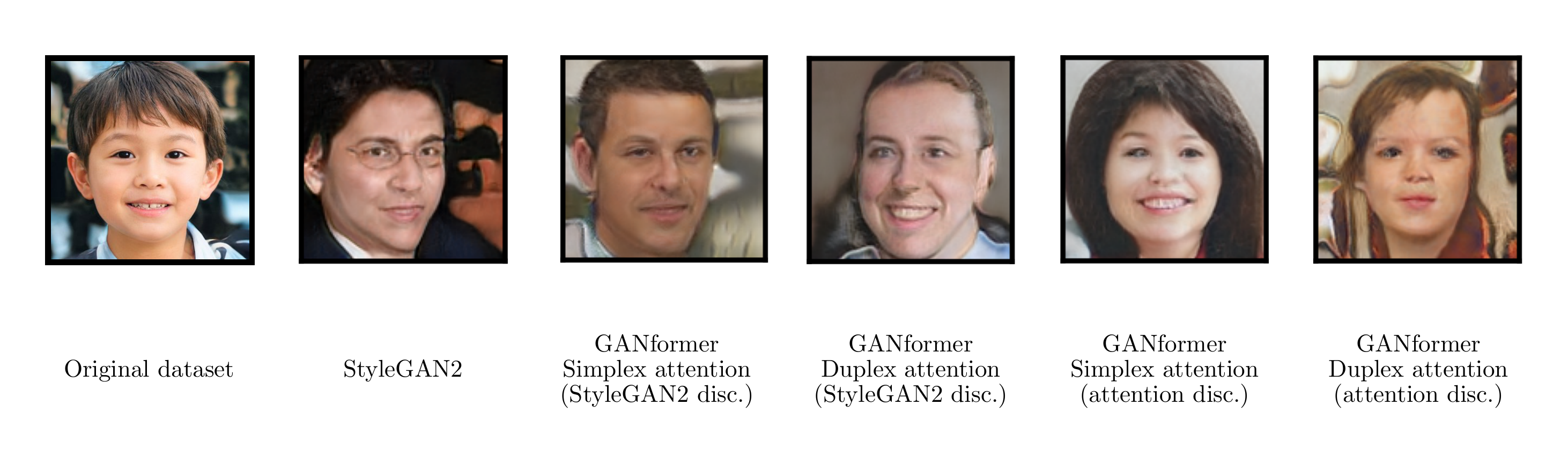}
	\caption{\textbf{Images generated using of the various models}.} 
	\label{fig:summaryImgs-ffhq}
\end{figure}
Figure~\ref{fig:summaryImgs} shows an image generated by each variation. Even for this dataset, the simplex GANformer implementation with StyleGAN2 is the model that performs better. 

Finally, to visualise our findings less empirically, as for the previous dataset, we have used random seeds to create latent inputs and shown the resulting images generated by the baseline StyleGAN2 and all four presented variations of the GANformer in Figures \ref{fig:random-ffhq} and \ref{fig:interpolation-ffhq} in Appendix \ref{sec:ffhq-results}.

\section{Discussion and conclusion}
%	Here you can express your judgments and draw your conclusions based on the  evidences 
%	produced on the previous sections.
%	
%	Try to summarise the achievements of your project and its limits, suggesting (when appropriate) 
%	possible extensions and future works.

In this project, we have discussed the replication of ``\emph{Generative adversarial Transformers}'' by \citet{hudson2021generative}.
%Contrary to our initial expectations, in our opinion, some of the claims made by the authors were 
%misleading or unproven in a scientific manner.
%Once we made this realisation, our methodology shifted from a pure reproduction of the results to a 
%search of undisclosed variations of the novel GANformer system in the published material trying to 
%obtain results that were comparable to the given ones.
As mentioned in Section~\ref{sec:comput_req}, our limited resources forced us to reduce the depth of our experiments compared to the authors'.
We had to shift from four datasets to two, size-reduced one: \citet{cartoonset}, significantly decreasing the time of execution and the complexity of our research.
For the same reasons, we have discarded four baselines: GAN, k-GAN, SAGAN and VQGAN in favour of just StyleGAN2.
The choice of StyleGAN2 as the baseline was prompted by a similarity of its execution to the novel GANformer, where parts of the author's code are a carbon copy of the StyleGAN2 implementation first proposed in \citet{karras2019style}.

In this study, we have first implemented a Google Colab Pro compatible version of the authors' code, enabling us to train and test StyleGAN2 and the two flavours of GANformer (Duplex and Simplex attention) in less than 54 hours in total.
As presented in Section~\ref{sec:results}, we could not achieve the improvements declared by the authors but instead found a decrease in time performance and quality using duplex attention.

Believing it was an error arising from our adaptation, we have analysed the pre-trained models provided and noticed what we think are discrepancies between the code and the methodology discussed in the paper.

In ``\emph{Generative Adversarial Transformers}'' by \citet{hudson2021generative}, the attention component is said to be placed both on the Generator and the discriminator sub-networks. At the same time, in all the pre-trained models provided, it seems never to be used on the discriminator.

%Proofreading the code, a BUG has been discovered by us, the optional attention flag, even if set to 
%True 
% when selecting a network structure, is never read during model preparations, resulting in models 
%that never 
% make use of attention in the discriminator phase.

%Moreover, to us it seemed that a discrepancy could be found in the Hyper-parameters settings.
Believing that all of the authors' experiments were performed on a different network than the one presented in the publication, we have tried to recreate the structure that could have yielded the claimed qualitative results.
To perform such a task, we have decomposed the GANformer network in two sections by keeping the presented Generator but substituting the hypothetically valid discriminator network with a vanilla StyleGAN discriminator.

Surprisingly, the StyleGAN/GANformer hybrid performed significantly better than the baseline. 

In conclusion, we have successfully reproduced the ``\emph{Generative Adversarial Transformers}'' by \citet{hudson2021generative} and found unexpected results.
We have then modified the proposed methodologies to obtain an image generation network that takes inspiration from the novel GANformer and adapts it to produce images that score significantly higher than the baseline over four quality metrics.

%\section{Authors contribution}
%\comment{review}{use the standard CRediT authorship contribution statement}
%All the authors analysed the original article and the code provided by the creators. 
%They took together all the important decisions, either regarding the organisation of the work and the 
%problems that have arisen.
%
%Stefano dealt with the dataset preparation and preprocessing phase. He also adapted the StyleGAN2 
%baseline for Colab and performed the training loop.
%
%Felix dealt with the given code and take charge of merging the StyleGAN2 and GANformer models, 
%adapting the code so that it reproduced carefully what was written in the paper.
%
%Stefano and Giorgia collaborated on the adaptation of the style transformer for StyleGAN2 and later 
%on the adaptation of the image visualisation for the GANformer, and in general with the generation 
%of images and videos. 
%
%Stefano and Felix worked together on the style mixing adaptation for the GANformer, and since only 
%they had available the GPU resources from Google Colab, they also handled with the models training.
%
%Stefano also concentrated on the code refactoring phase, mainly cleaning the code related to the 
%discriminator GANformer.
%
%Giorgia focused on the main conceptual ideas behind the original work, and performed a theoretical 
%research on the topic and related works. 
%Later she drafted this manuscript, with inputs from the other authors which provided critical 
%feedback on it. 
%
%All the authors discussed and interpret the results and finally contribute to the final version of this 
%article.

\clearpage
\bibliography{bibliography}
\bibliographystyle{unsrtnat}

\clearpage
\appendix
\section{Background and methodology}\label{sec:appendix}
	
\subsection{Background} \label{subsec:app_background}
As aforementioned in Section~\ref{sec:gan}, GANs \cite{goodfellow2014generative}, are deep-learning-based generative models composed by two main neural networks: the \textit{generator} ${G(z)}$, which take as input a sample from the latent space $z$, or a vector drawn randomly from a Gaussian distribution, and use it to generate new plausible examples in the problem domain — images in our case —, and the \textit{discriminator} ${D(x)}$, which takes an example from the domain as input, and predicts a binary class label which classifies the examples as real (coming from the training dataset) or fake (generated by $G$). 
	
The two models are trained together: the discriminator $D$ estimates the probability that sample $x$ is generated by $G$ or is a real sample and aims at maximising the probability of assigning the correct label to both real and fake samples. In contrast, the generator $G$ is trained to maximise the likelihood of the discriminator $D$ making a mistake, so it aims to minimise $\log(1-D(G(z))))$.
	
Combining the two objectives for $G(z)$ and $D(x)$ we get the \textit{GAN min-max game} with the value function $V(G,D)$:
\begin{equation}
    \label{e:minmaxgame}
    \min_G \max_D V(G,D) = 
    \mathbb{E}_{x \sim p*(x)} [\log D(x)] + \mathbb E _{z \sim p_z(z)} [\log (1-D(G(z))))]
\end{equation}
	
GANs typically work with images, as in our case. For this reason, both the generator and discriminator models use Convolutional Neural Networks (CNNs).
\begin{figure*}[htb]				
    \centering
    \includegraphics[width=.65\linewidth]{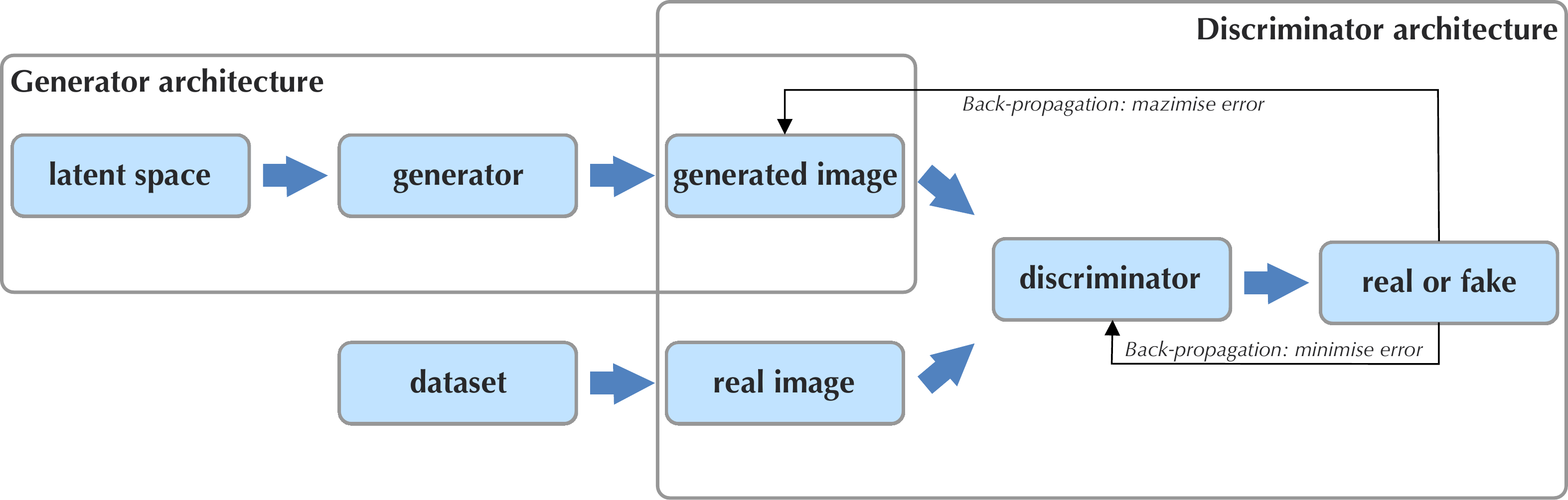}
    \caption{\textbf{GAN architecture.} Visualisation of the two components of a GAN: generator and discriminator.}
    \label{fig:gan}
\end{figure*}

\begin{figure*}[!h]				
    \centering
    \includegraphics[width=.5\linewidth]{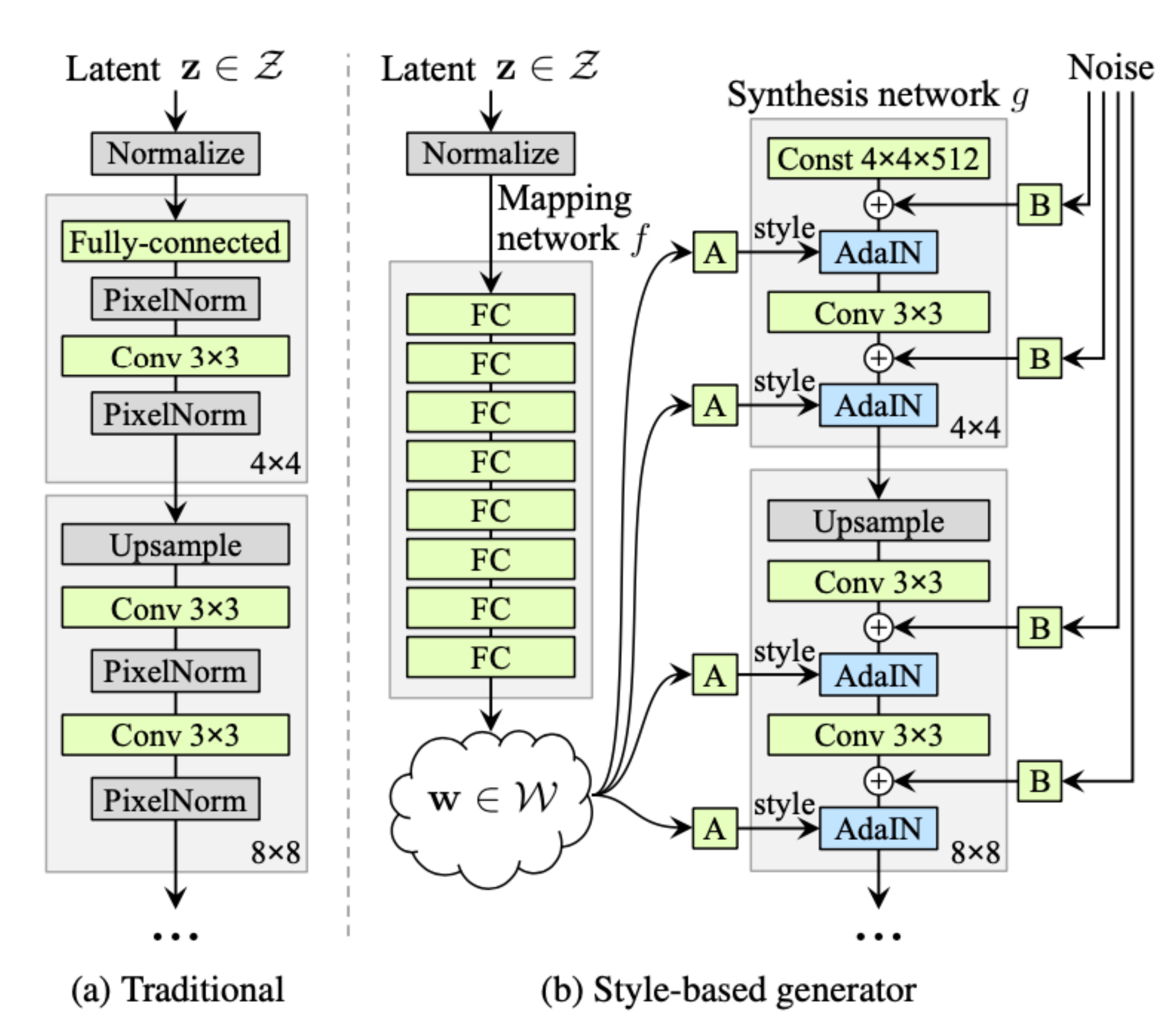}
    \caption{\textbf{StyleGAN architecture.} Re-design of the traditional GAN generator into a style-based generator.}
    \label{fig:StyleGAN}
\end{figure*}

In this paper, we exploited StyleGAN2 architecture, the second version of the StyleGAN model, introduced in Section~\ref{sec:StyleGAN}, a re-design of the GANs generator architecture.
The StyleGAN aims to control the image synthesis process \cite{karras2019style}. The architecture is illustrated in Figure~\ref{fig:StyleGAN}.

The StyleGAN approach departs from the design of a standard generator network consisting of a multi-layer CNN.
To start the up-sampling process, a traditional GAN generator only feeds the latent code $z$ through the input layer.
\citet{karras2019style} introduce a feed-forward mapping network $f : Z \rightarrow W$ which processes the latent code $z$ in the input latent space $Z$, and output an intermediate latent vector $w \in W$. 
$w$, in turn, interacts directly with each convolution through the synthesis network  $g$. In particular, the \textit{Adaptive Instance Normalisation (AdaIN)} \cite{huang2017arbitrary} aligns the mean and variance of the content features with those of the style features, meaning that can globally control these parameters and so the strength of image features at different scales. 

The \textit{AdaIN} operation is defined as:
\begin{equation}
    \label{e:adain}
    \mathsf{AdaIN}(x, y) = \sigma(y) \bigg(\frac{x - \mu(x)}{\sigma (x)} \bigg) + \mu (y) \mbox{,}
\end{equation}
where $x$ is the content input, $y$ is a style input, and the channel-wise mean and variance of $x$ are aligned to match those of $y$.

\citet{karras2019style} provided the generator with a direct means to generate stochastic details by introducing direct Gaussian noise inputs after each convolution: an automatic and unsupervised separation of high-level attributes (e.g., pose, identity) from stochastic variation (e.g., freckles, hair) is obtained in the generated images, and intuitive scale-specific mixing and interpolation operations are enabled.

\begin{figure*}[htb]				
    \centering
    \includegraphics[width=.9\linewidth]{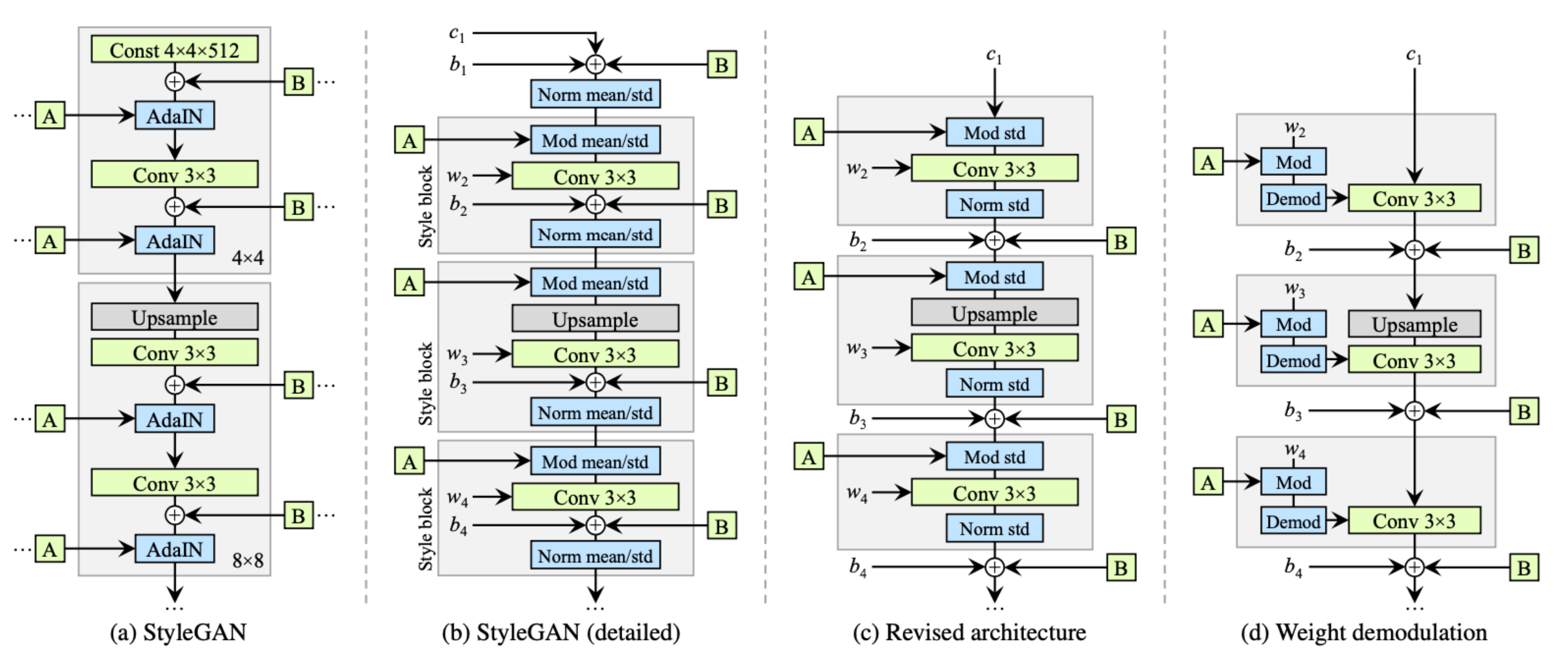}
    \caption{\textbf{StyleGAN2 architecture.} Evolution from a StyleGAN to a StyleGAN2 synthesis network.}
    \label{fig:StyleGAN2}
\end{figure*}

StyleGAN2 \cite{karras2020analyzing} is a revisiting of the architecture of the StyleGAN synthesis network. Figure~\ref{fig:StyleGAN2} shows the changes made to the original architecture up to the final network.

In particular, some redundant operations have been removed from the original StyleGAN architecture.
Bias and noise are operations which might have conflicting interests, and their application within the style block caused their relative impact to be inversely proportional to the current style's magnitudes. For this reason, they have been moved outside the style block and separated, allowing them to obtain more predictable results.
Furthermore, the mean is no longer necessary after this change, but it is sufficient for the normalisation and modulation to operate on the standard deviation alone.
Moreover, the application of bias, noise, and normalisation to the constant input has been safely removed. 

When the style block consists of a modulation, a convolution, and a normalisation operation, it is possible to restructure the AdaIN has a demodulation operation, which is applied to the weights associated with each convolution layer. 
AdaIN uses different scale and shift parameters to align other areas of $w$ — the activations of intermediate activations — with different regions of the feature map (either within each feature map or via grouping features channel-wise by spatial location). At the same time, weight demodulation takes the scale and shift parameters out of a sequential computation path, instead baking scaling into the parameters of convolutional layers.

%	The \textit{mapping network}, which consists of a series of feed-forward layers, aims to convert the sampled latents from a normal distribution $z$ to the intermediate space $w$. The $k$ latent components either are mapped independently or interact with each other through self-attention.
%	The \textit{synthesis network} is composed by multiple layers of convolution where the image's features begin from a small constant/sampled grid and up-sampling until reaching the desired resolution.
%	After each convolution, the image features are modulated by the intermediate latent vectors $w$, meaning that their variance and bias are controlled. 

Also, transformers are an influential deep-learning model exploited in this work and introduced in Section~\ref{sec:transformer}. 	
An \textit{attention function} \cite{vaswani2017attention} can be described as mapping a query and a set of key-value pairs to an output, where the query, keys, values, and output are all vectors. The output is computed as a weighted sum of the values, where a compatibility function of the query with the corresponding key calculates the weight assigned to each value.

The input consists of queries and keys of dimension $d_k$ and size values $d_v$. The dot products of the query with all keys are computed, each divided by $\sqrt{d_k}$ and then a softmax function is applied to obtain the weights on the values.
In practice, the attention function is computed simultaneously on queries packed into a matrix $Q$. The keys and values are also packed into matrices $K$ and $V$. 
\begin{equation}
    \label{eqn:attention}
    \mathsf{Attention}(Q, K, V) = \mathsf{softmax} \big( \frac{QK^T}{\sqrt{d_k}}\big)V
\end{equation}
% This is a \textit{dot-product attention} with a scaling factor of $\sqrt{1}$. 
Instead of performing a single attention function, transformers use multiple self-attentions, called \textit{multi-head attention}, allowing the model to jointly attend to information from different representation subspaces at different positions, learning attention relationship independently \cite{vaswani2017attention}.
As mentioned in	Section~\ref{sec:transformer}, \citet{vaswani2017attention} used multiple multi-head attention, defined as follows.
\begin{equation}
    \label{eqn:multihead}
    {\mathsf{MultiHead}(Q, K, V) = \mathsf{Concat}(\mathsf{head}_1 \dots, 
        \mathsf{head}_h) W^O }
    \mbox{,}
\end{equation}
where $ \mathsf{head}_i = \mathsf{Attention}(QW_i^Q, KW_i^K , VW_i^V)$, and the projections are parameter matrices $W_i^Q \in \mathbb{R}^{d_{\text{model}}\times d_k}$, $W_i^K \in \mathbb{R}^{d_{\text{model}}\times d_k}$, $W_i^V \in \mathbb{R}^{d_{\text{model}}\times d_v}$ and $W^O \in \mathbb{R}^{hd_v \times d_{\text{model}}}$.

As reported by \cite{vaswani2017attention}, the transformer uses multi-head attention in three different ways:
\begin{enumerate}
    \item In "encoder-decoder attention" layers, the queries come from the previous decoder layer, and the memory keys and values come from the output of the encoder. This allows every position in the decoder to attend to all positions in the input sequence.
    \item The encoder contains self-attention layers in which all of the keys, values and queries come from the output of the previous layer in the encoder. 
    Each position in the encoder can attend to all positions in the previous layer of the encoder.
    \item Similarly, self-attention layers in the decoder allow each position in the decoder to attend to all positions up to and including that position. We must prevent leftward information flow in the decoder to preserve the auto-regressive property. We implement this inside of scaled dot-product attention by masking out (setting to $-\infty$) all values in the input of the softmax, which correspond to illegal connections. 
\end{enumerate}

%	At each step, the model is auto-regressive, consuming the previously generated symbols as 
%	additional input when generating the next.
%		\begin{enumerate*}
%			\item[(1)] a \textit{multi-head self-attention mechanism} working in parallel, and 
%			\item[(2)] a \textit{position-wise fully connected feed-forward neural network}, with 
%residual 
%			connections followed by normalisation.
%	\end{enumerate*}
%	Moreover, the decoder self-attention sub-layer has a mask which prevents positions from 
%	attending 
%	to subsequent positions (so positional encoding provided as additional input to bottom 
%layer). 
%	This, combined with fact that the output embeddings are offset by one position, ensures 
%that 
%	the predictions for position $i$ can depend only on the known outputs at positions less 
%than $i$.
% 	In addition, the decoder has a third sub-layer, which performs \textit{multi-head attention} 
%over 
%	the output of the encoder stack. 

\subsection{Generative Adversarial Transformers in depth} \label{subsec:app_methodology}
In this section, we dive into the details of Generative Adversarial Transformers \cite{hudson2021generative}, introduced in Section~\ref{sec:ganformer}. 

The BERT \textit{transformer network} \cite{devlin2019bert} interleaves \textit{multi-head self-attention} and \textit{feed-forward layers}. Each pair of self-attention and feed-forward layers is intended as a \textit{transformer layer}. Hence, a transformer is a stack of several such layers. 

The \textit{self-attention layer} considers all pairwise relations among the input elements to update each element by attending to all the others. 
The \textit{bipartite transformer} generalises this formulation, featuring a bipartite graph between two groups of variables instead — in the GAN case, latent and image features. 

In particular, the attention layers are added between the convolutional layers of the generator and discriminator.

Instead of controlling the style of all features globally, the new attention layer is used to perform \textit{adaptive region-wise modulation}. The latent vector $z$ is split into $k$ components, $z = [z_1, \dots, z_k ]$ and, as in StyleGAN \cite{karras2019style}, pass each of them through a shared mapping network, obtaining a corresponding set of intermediate latent variables $Y = [y_1, \dots,y_k ]$. 

During synthesis, after each CNN layer in the generator, the feature map $X$ and latents $Y$ play the roles of the two-element groups, mediating their interaction through our new attention layer (either simplex or duplex). 
This setting thus allows for a flexible and dynamic style modulation at the region level. 

Since soft attention tends to group elements based on proximity and content similarity, the transformer architecture naturally fits into the generative task. It proves helpful in the visual domain, allowing the model to exercise finer control in modulating local semantic regions. This capability turns out to be especially useful in modelling highly-structured scenes.

For the discriminator, attention is applied after every convolution using trained embeddings to initialise the aggregator variables $Y$, which may intuitively represent background knowledge the model learns about the task. At the last layer, these variables $Y$ are concatenated to the final feature map $X$ to predict the identity of the image source. 
This structure empowers the discriminator with the capacity to model long-range dependencies, which can aid it in assessing image fidelity, allowing it to acquire a more holistic understanding of the visual modality.
%	\textcolor{blue}{
%	\begin{itemize}
%		\item Compositional Latent Space with multiple variables that coordinate through 
%attention to producing the image cooperatively, matching the inherent compositionality of natural scenes.
%		\item Bipartite Structure that balances between expressiveness and efficiency, modelling long-range dependencies while maintaining linear computational costs.
%		\item Bidirectional Interaction between the latents and the visual features allows the refinement and interpretation of each in light of the other.
%		\item Multiplicative Integration rule to impact the features' visual style more flexibly, akin to StyleGAN but in contrast to the transformer network.
%	\end{itemize}
%}

\section{Google Cartoon Set results} \label{sec:cartoon-results}
The dataset used in this work is the Google Cartoon Set \cite{cartoonset} introduced in Section~\ref{sec:dataset}, containing 10k 2D cartoon avatars. These images comprise 16 components that vary in 10 artwork attributes, 4 colour attributes, and 4 proportion attributes (see Table~\ref{tab:dataset}). 

\begin{table}[htb]
	\centering
	\caption{Attributes of the Cartoon Set.}
	\label{tab:dataset}
	\vspace{3mm}
	\small
	\begin{tabularx}{\textwidth}{ll|r|X}
		&& \textbf{\# Variants} & \textbf{Description}                              \\
		\toprule
		\multirow{10}*{\textbf{Artwork}} 	&	\texttt{chin\_length}           & 3           & Length of 
		chin 
		(below 	mouth region)      \\
		&	\texttt{eye\_angle}             & 3           & Tilt of the eye inwards or outwards      \\
		&	\texttt{eye\_lashes}            & 2           & Whether or not eyelashes are visible     \\
		&	\texttt{eye\_lid}               & 2           & Appearance of the eyelids      	\\
		&	\texttt{eyebrow\_shape}        & 14          & Shape of eyebrows        \\
		&	\texttt{eyebrow\_weight}        & 2           & Line weight of eyebrows           \\
		&	\texttt{face\_shape}            & 7           & Overall shape of the face                \\
		&	\texttt{facial\_hair}           & 15          & Type of facial hair (type 14 is no hair) \\
		&	\texttt{glasses}                & 12          & Type of glasses (type 11 is no glasses)  \\
		&	\texttt{hair}                   & 111         & Type of head hair                        \\
		\midrule
		\multirow{4}*{\textbf{Colors}} &	\texttt{eye\_color}    & 5 & Color of the eye 
		irises           \\
		&	\texttt{face\_color}            & 11          & Color of the face skin                   \\
		&	\texttt{glasses\_color}         & 7           & Color of the glasses, if present         \\
		&	\texttt{hair\_color}        & 10     & Color of the hair, facial hair, and eyebrows      \\
		\midrule
		\multirow{4}*{\textbf{Proportions}} &	\texttt{eye\_eyebrow\_distance} & 3           & 
		Distance 
		between the eye and eyebrows    \\
		&	\texttt{eye\_slant}             & 3           & Similar to eye\_angle, but rotates the eye and 
		does 
		not change artwork  \\
		&	\texttt{eyebrow\_thickness}     & 4           & Vertical scaling of the eyebrows         \\
		&	\texttt{eyebrow\_width}         & 3           & Horizontal scaling of the eyebrows            \\
		\bottomrule                         
	\end{tabularx}
\end{table}

\begin{figure}[htpb]
	\centering
	\begin{subfigure}{\linewidth}
		\includegraphics[width=\linewidth]{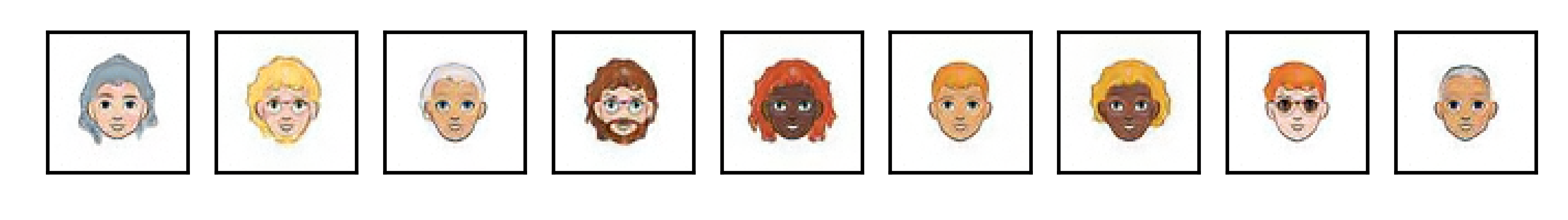}
		\vspace{-7mm}
		\caption{StyleGAN2.} 
	\end{subfigure}
	\begin{subfigure}{\linewidth}
		\includegraphics[width=\linewidth]{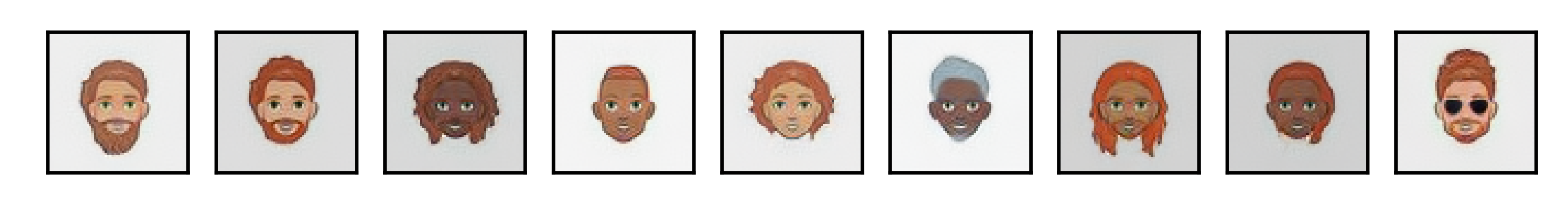}
		\vspace{-7mm}
		\caption{GANformer with Simplex attention and vanilla StyleGAN2 discriminator.}
	\end{subfigure}
	\begin{subfigure}{\linewidth}
		\includegraphics[width=\linewidth]{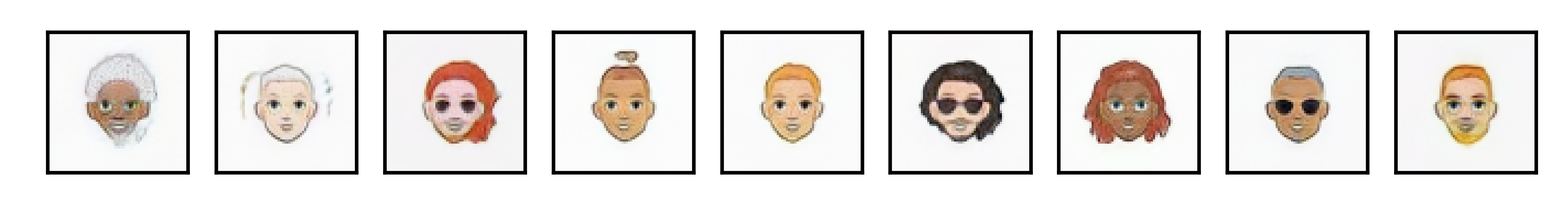}
		\vspace{-7mm}
		\caption{GANformer with Simplex attention.}
	\end{subfigure}
	\begin{subfigure}{\linewidth}
		\includegraphics[width=\linewidth]{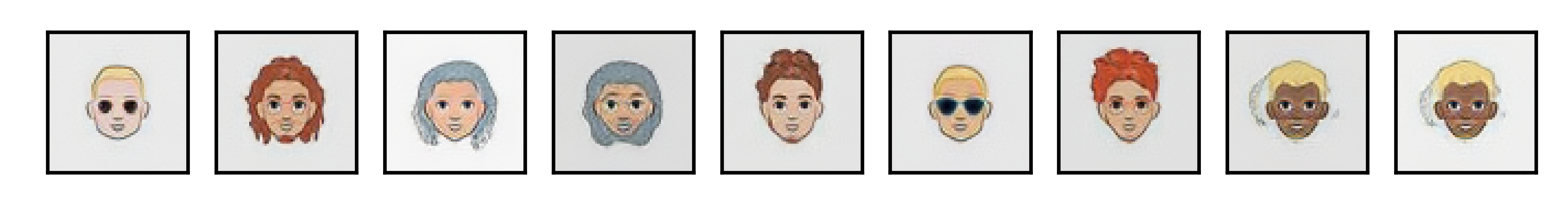}
		\vspace{-7mm}
		\caption{GANformer with Duplex attention and vanilla StyleGAN2 discriminator.}
	\end{subfigure}
	\begin{subfigure}{\linewidth}
		\includegraphics[width=\linewidth]{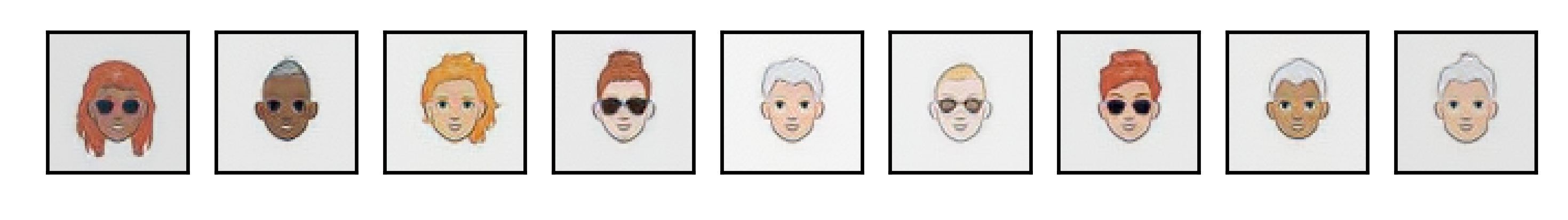}
		\vspace{-7mm}
		\caption{GANformer with Duplex attention.}
	\end{subfigure}
	\vspace{3mm}
	\caption{\textbf{Visualisation of 9 images generated with the various models}.}\label{fig:random}
\end{figure}

\begin{figure}[htpb]
	\centering
	\begin{subfigure}{\linewidth}
		\includegraphics[width=\linewidth]{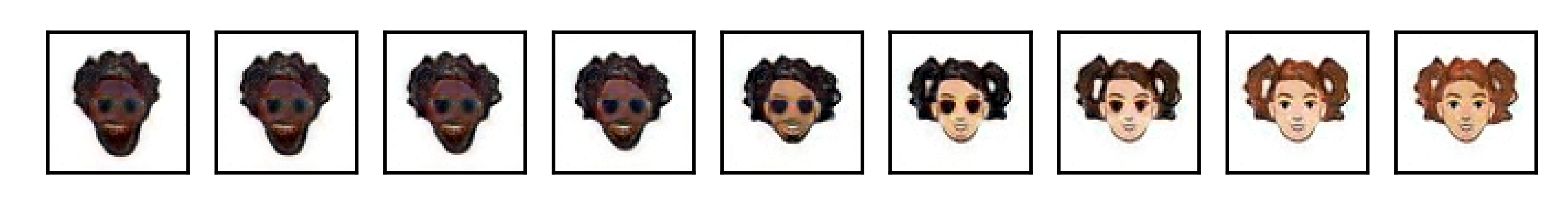}
		\vspace{-7mm}
		\caption{StyleGAN2.} 
	\end{subfigure}
	\begin{subfigure}{\linewidth}
		\includegraphics[width=\linewidth]{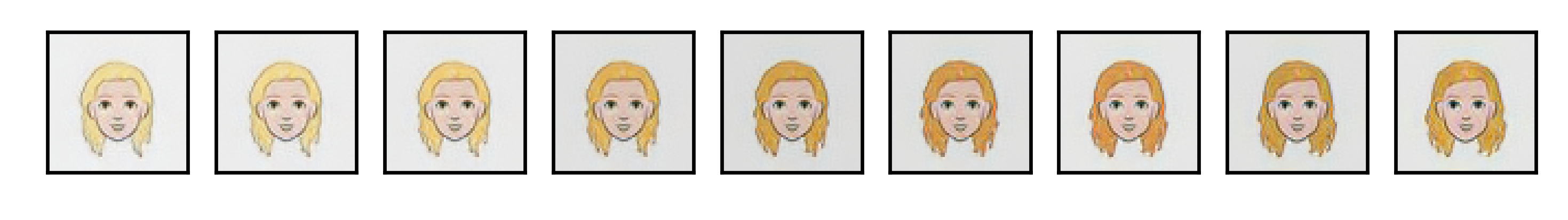}
		\vspace{-7mm}
		\caption{GANformer with Simplex attention and vanilla StyleGAN2 discriminator.}
	\end{subfigure}
	\begin{subfigure}{\linewidth}
		\includegraphics[width=\linewidth]{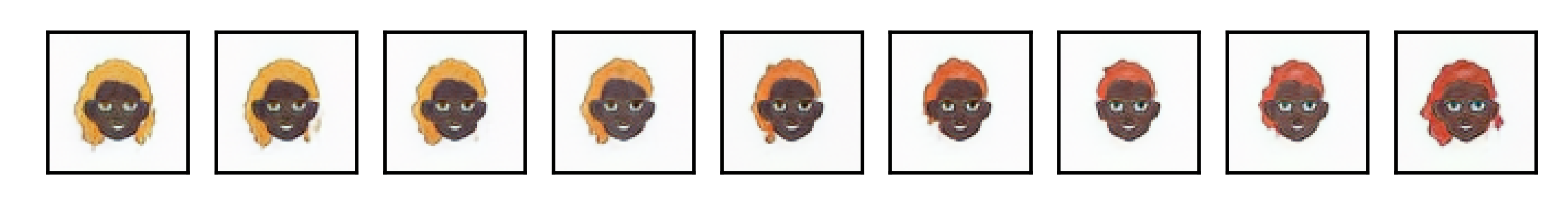}
		\vspace{-7mm}
		\caption{GANformer with Simplex attention.}
	\end{subfigure}
	\begin{subfigure}{\linewidth}
		\includegraphics[width=\linewidth]{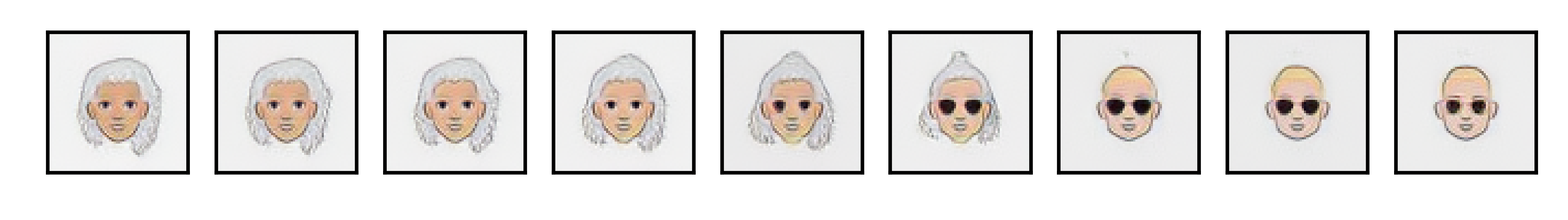}
		\vspace{-7mm}
		\caption{GANformer with Duplex attention and vanilla StyleGAN2 discriminator.}
	\end{subfigure}
	\begin{subfigure}{\linewidth}
		\includegraphics[width=\linewidth]{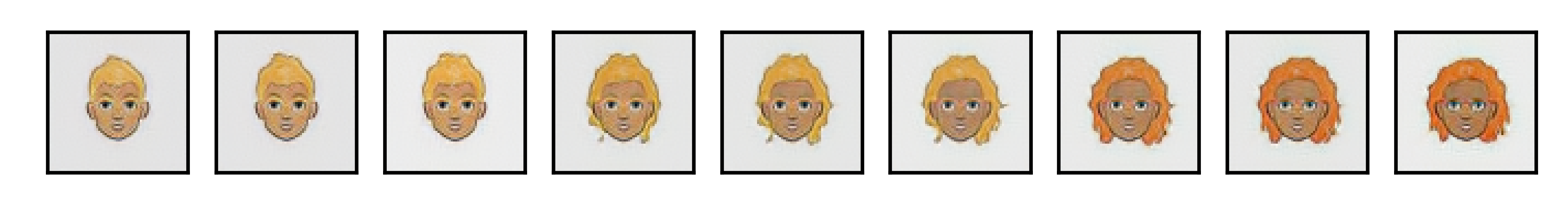}
		\vspace{-7mm}
		\caption{GANformer with Duplex attention.}
	\end{subfigure}
	\vspace{3mm}
	\caption{\textbf{Simple $\mathbf{z}$ interpolation using of the various models}.} 
	\label{fig:interpolation}
\end{figure}

\clearpage 
\section{FFHQ dataset results} \label{sec:ffhq-results}
\begin{figure}[htpb]
	\centering
	\begin{subfigure}{\linewidth}
		\includegraphics[width=\linewidth]{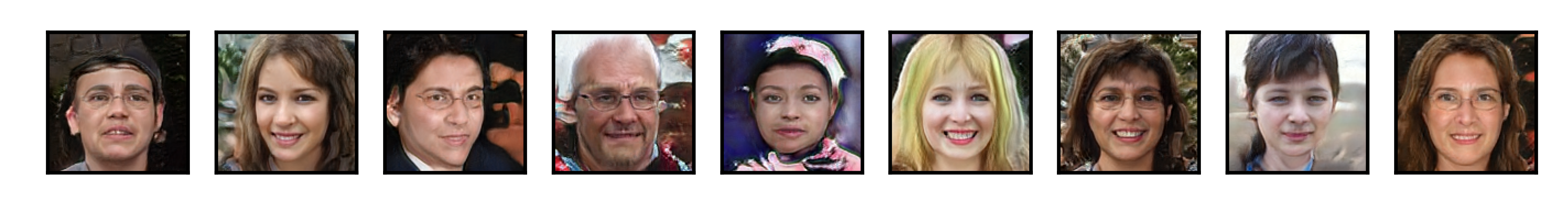}
		\vspace{-7mm}
		\caption{StyleGAN2.} 
	\end{subfigure}
	\begin{subfigure}{\linewidth}
		\includegraphics[width=\linewidth]{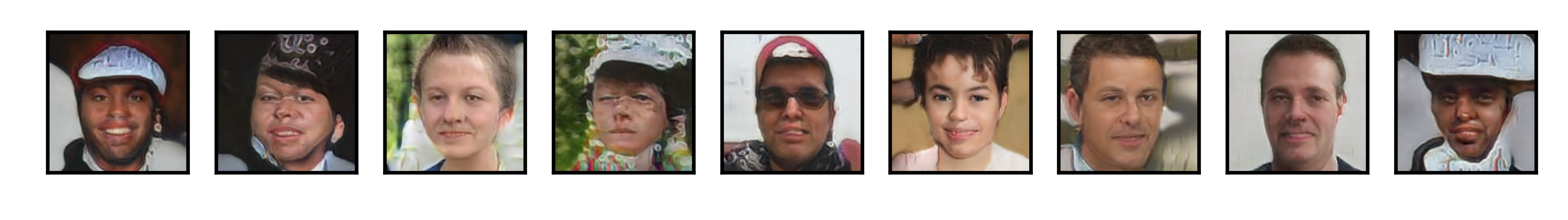}
		\vspace{-7mm}
		\caption{GANformer with Simplex attention and vanilla StyleGAN2 discriminator.}
	\end{subfigure}
	\begin{subfigure}{\linewidth}
		\includegraphics[width=\linewidth]{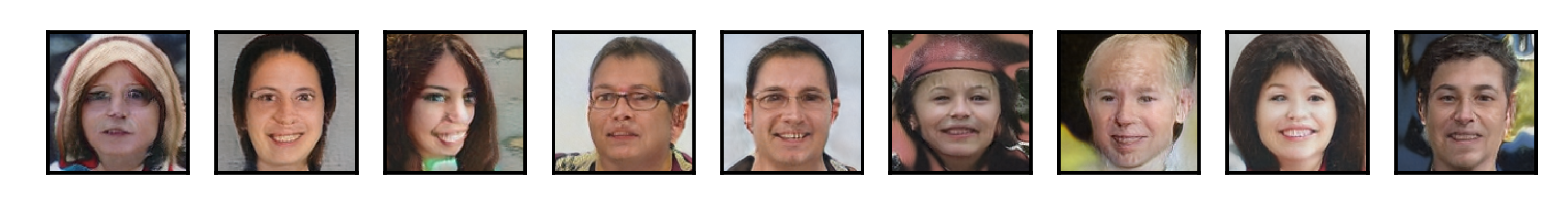}
		\vspace{-7mm}
		\caption{GANformer with Simplex attention.}
	\end{subfigure}
	\begin{subfigure}{\linewidth}
		\includegraphics[width=\linewidth]{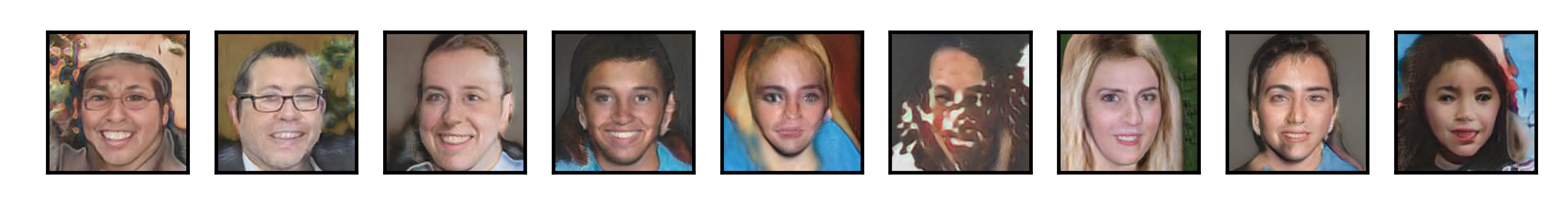}
		\vspace{-7mm}
		\caption{GANformer with Duplex attention and vanilla StyleGAN2 discriminator.}
	\end{subfigure}
	\begin{subfigure}{\linewidth}
		\includegraphics[width=\linewidth]{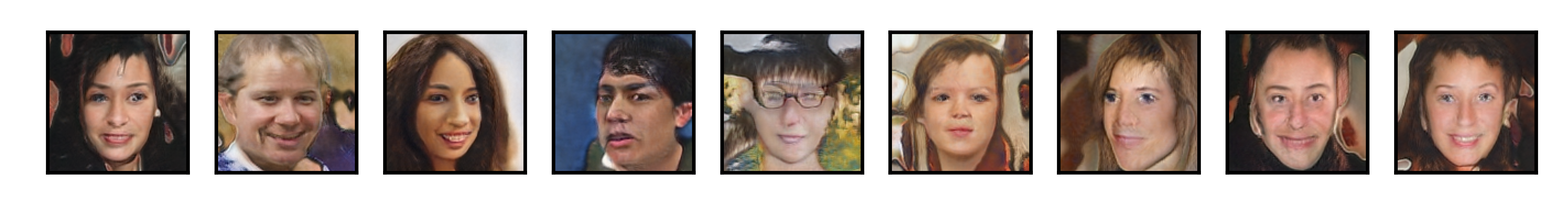}
		\vspace{-7mm}
		\caption{GANformer with Duplex attention.}
	\end{subfigure}
	\vspace{3mm}
	\caption{\textbf{Visualisation of 9 images generated with the various models}.}\label{fig:random-ffhq}
\end{figure}

\begin{figure}[htpb]
	\centering
	\begin{subfigure}{\linewidth}
		\includegraphics[width=\linewidth]{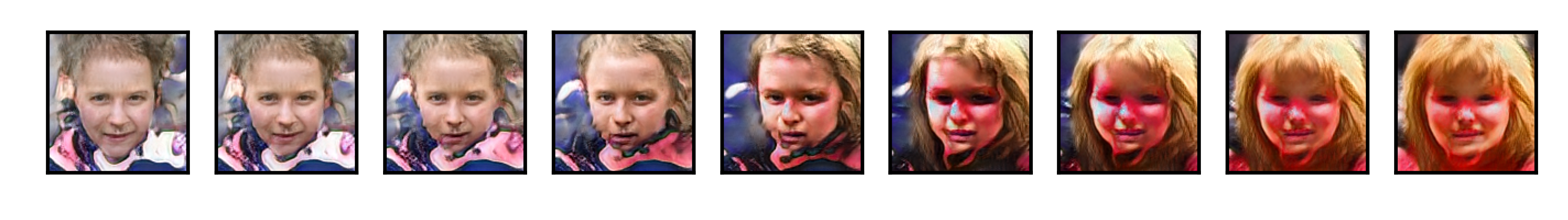}
		\vspace{-7mm}
		\caption{StyleGAN2.} 
	\end{subfigure}
	\begin{subfigure}{\linewidth}
		\includegraphics[width=\linewidth]{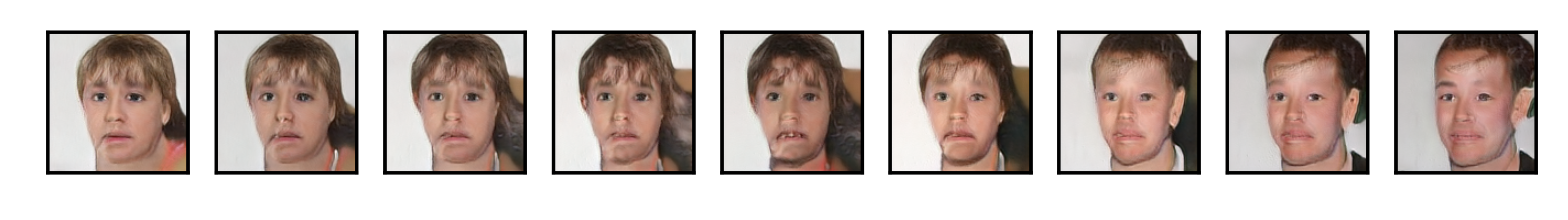}
		\vspace{-7mm}
		\caption{GANformer with Simplex attention and vanilla StyleGAN2 discriminator.}
	\end{subfigure}
	\begin{subfigure}{\linewidth}
		\includegraphics[width=\linewidth]{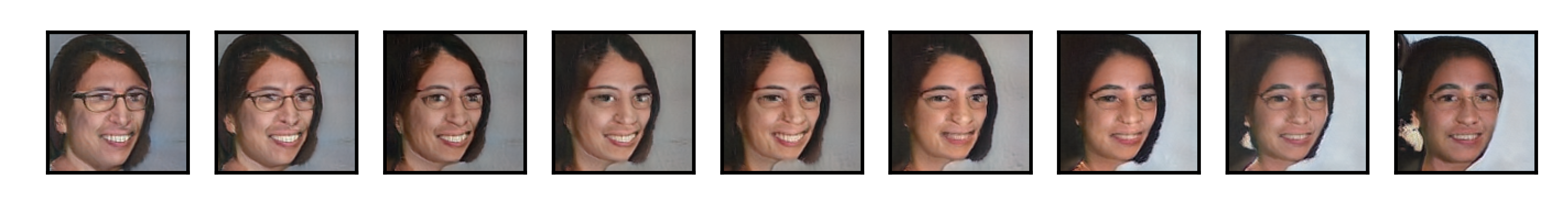}
		\vspace{-7mm}
		\caption{GANformer with Simplex attention.}
	\end{subfigure}
	\begin{subfigure}{\linewidth}
		\includegraphics[width=\linewidth]{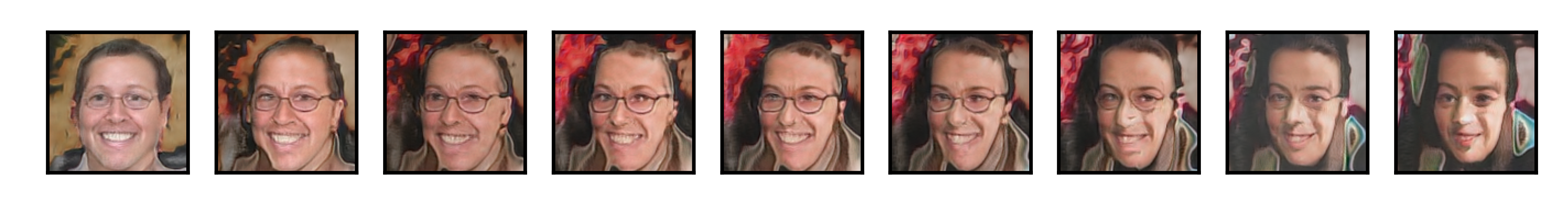}
		\vspace{-7mm}
		\caption{GANformer with Duplex attention and vanilla StyleGAN2 discriminator.}
	\end{subfigure}
	\begin{subfigure}{\linewidth}
		\includegraphics[width=\linewidth]{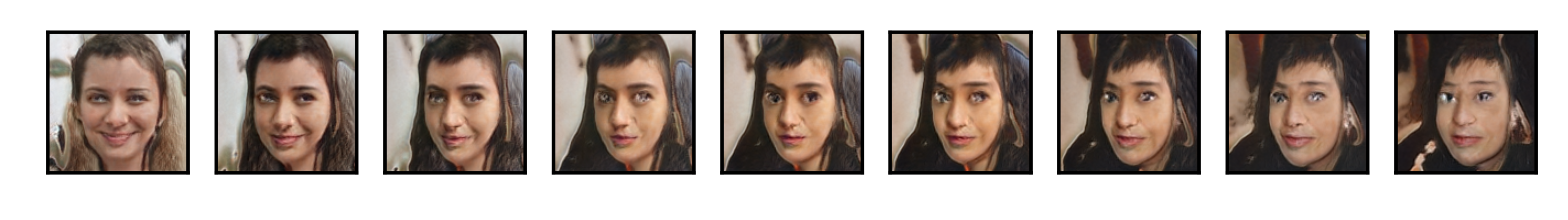}
		\vspace{-7mm}
		\caption{GANformer with Duplex attention.}
	\end{subfigure}
	\vspace{3mm}
	\caption{\textbf{Simple $\mathbf{z}$ interpolation using of the various models}.} 
	\label{fig:interpolation-ffhq}
\end{figure}

\end{document}